# Normalized Orthography for Tunisian Arabic

Houcemeddine Turki[1][0000−0003−3492−2014], Kawthar Ellouze[2], Hager Ben Ammar[2], Mohamed Ali Hadj Taieb[1][0000−0002−2786−8913], Imed Adel[3], Mohamed Ben Aouicha[1][0000−0002−2277−5814], Pier Luigi Farri[4], and Abderrezak Bennour[1]

[1] Data Engineering and Semantics Research Unit, University of Sfax, Sfax, Tunisia
turkiabdelwaheb@hotmail.fr
[2] Association Derja, Tunis, Tunisia
[3] University of Sousse, Sousse, Tunisia
[4] VBScuola, Rome, Italy

**Abstract.** Tunisian Arabic (ISO 693-3: aeb) is a distinct variety native to Tunisia, derived from Arabic and enriched by various historical influences. This research introduces the "Normalized Orthography for Tunisian Arabic" (NOTA), an adaptation of CODA* guidelines for transcribing Tunisian Arabic using Arabic script. The aim is to enhance language resource development by ensuring user-friendliness and consistency. The updated standard addresses challenges in accurately representing Tunisian phonology and morphology, correcting issues from transcriptions based on Modern Standard Arabic.

## 1 Introduction

Tunisian (ISO 693-3: aeb) is a linguistic variety native to Tunisia [12], emerging from Arabic with influences from Latin, Berber, Italian, French, Spanish, Turkish, Mediterranean Lingua Franca, and possibly Neo-Punic [35]. These influences shaped Tunisian into a distinct entity, separate from Classical Arabic [35]. Adoption of the Arabic script in the 19th century facilitated systematic recording and study of the language. This transition was triggered in the 1930s as part of Nationalist efforts to promote independence from France and improve literacy among Tunisians [28]. Literary production in Tunisian, led by intellectuals like Ali Douagi, gained momentum [28]. The Tunisian pro-independence party Neo-Destour collaborated with linguist William Marçais to describe dialects and create text corpora for Tunisian [19, pp. 216-250]. This initiative waned following post-independence Arabization policies but resurfaced after the Tunisian revolution [28].

These efforts, drawing inspiration from Modern Standard Arabic orthography to transcribe Tunisian [39], were formalized as guidelines by Maamouri et al. (2004) [23] and Makhoul et al. (2005) [5]. Habash et al. (2012) [15] expanded this into CODA, with adaptations for specific Arabic languages, includ-ing Tunisian [45,43] and Maghrebi Arabic [39]. These adaptations later amalgamated into CODA*, facilitating comparative analyses [2], part-of-speech tagging [1], sentiment analysis [33], morphological modeling [42], and corpus creation [3]. Despite consensus among computational linguists, CODA* has not yet seen widespread adoption or evaluation by the general audience, motivating further exploration. CODA* Guidelines, designed for transcribing Arabic dialects, may pose limitations for speakers unfamiliar with Modern Standard Arabic, resulting in conflicting transcriptions and impeding consistent morphological patterns for Tunisian Arabic. Trials to solve these issues by developing statistical and computational approaches for normalizing the transcription of Tunisian using the CODA* Guidelines have been proposed by Besdouri et al. (2021) [8] and Mekki et al. (2022) [29]. However, limited efforts have been made to develop an orthography normalization for Tunisian using the principles of Applied Linguistics.

In this research paper, we introduce our adaptation of the CODA* guidelines for transcribing Tunisian Arabic using the diacritized or undiacritized Arabic script. This will allow us to benefit from the achievements of the CODA* Guidelines instead of developing a new standard from scratch. We present the resulting "Normalized Orthography for Tunisian Arabic" (NOTA), prioritizing user-friendliness and consistency. Our paper is structured as follows: We provide a linguistic overview of Tunisian (Section 2), specify the CODA* Guidelines for Arabic Languages (Section 3), explain our approach to developing NOTA (Section 4), describe our guidelines for transcribing Tunisian using the Arabic script (Section 5), explain possible applications of the NOTA Guidelines related to computational linguistics (Section 6), and conclude with future research directions (Section 7).

## 2   Linguistic Overview of Tunisian

Tunisian Arabic, a unique variant of Arabic, is the primary spoken language in Tunisia and is widely used in media, including music, television, and radio [12]. It shares features with other Maghreb languages [9]. Notable aspects include conservative consonant sounds, the urban use of "إنتِي [ɪnti]" (you), and no gender distinctions in the second person verb forms [12]. Unlike Modern Standard Arabic (MSA), Tunisian Arabic lacks an indicative prefix, blurring indicative and subjunctive moods [12]. It employs a progressive aspect using "قاعد [qɛːʕɪd]" (to remain doing something) and "في [fi]" (in) in transitive clauses [34]. The future tense uses prefixes like "باش [bɛːʃ]" or "بِش [bəʃ]" combined with verbs [12].

Tunisian Arabic follows a Subject-Verb-Object (SVO) order and generally acts as a Null-subject language [12]. It shows more agglutinative structures compared to Standard Arabic and other varieties, influenced by Turkish during the 17[th] century [13]. An example is "شنعمل [ʃnaʕmɪl]" (What should I do), a fusion of "آش [ɛːʃ]" (What) and "نعمل [naʕmɪl]" (I do). Dialects of Tunisian Arabic include Northwestern, Southwestern, Tunis, Sahel, Sfax, Judeo-Tunisian, and Southeastern. These mostly belong to the Hilalian group, except for the Sfax and Judeo-Tunisian variants [12]. While the voiceless uvular stop [q] is used in Tunisian, Southeastern, Northwestern, and Southwestern varieties use the voiced velar plosive [g] [12]. Some words consistently maintain the [g] sound, especially in agriculture, such as "بڨرة [bægra]" (cow), where /g/ derives from an originally Arabic [q], and "دڨلة [dɪglæ]" (dates) [4].

Phonologically, Tunisian Arabic is distinct from MSA. It predominantly retains interdental fricatives, except in the Sahel dialect [32], and merges /dˤ/ ض with /ðˤ/ ظ [20]. It includes a wide range of consonant phonemes, with emphatic consonants being relatively rare [20]. Non-Arabic words sometimes include /p/ and /v/, usually substituted with /b/ [20]. The glottal stop /ʔ/ is generally omitted but present in formal registers [20]. Shadda "gemination" frequently occurs, extending to final consonants in closed syllables [20]. Tunisian Arabic vowels include /a/, /ɛ/, /i/, and /u/, with vowel length being crucial [20]. Stress placement depends on syllable structure, favoring the ultimate syllable if doubly closed or the penultimate syllable otherwise [20]. Various phonological processes highlight its phonetic diversity and divergence from MSA [20].

The lexicon of Tunisian Arabic blends native, Latin, and Berber words with borrowed vocabulary from Italian, Spanish, French, and Turkish [38]. For example, "تريسيتي [triːsiːti]" (electricity) is borrowed from French, and "برتمان [bortmɛːn]" (flat) and "بياسة [bjɛːsæ]" (coin) have French roots. Turkish influence appears in words like "ڨاوري [gɛːwri]" (foreigner) and the occupation suffix "جي /ʒi/" [38]. Differences between Tunisian and Standard Arabic often result from shifts in the meanings of Arabic roots. For example, "خدم [xdɪm]" means to serve in Standard Arabic but to work in Tunisian, while "عمل [ʕmɪl]" means to work in Standard Arabic but has a broader sense of to do in Tunisian [6]. Word fusion is a significant mechanism in Tunisian Arabic, especially in question words, such as "اش [ɛːʃ]" or "ش [ʃ]" [12]. This feature is absent in MSA, Levantine, and Gulf Arabic [12].

## 3   CODA* Guidelines

CODA* is a conventional orthography for Dialectal Arabic (DA), designed primarily for computational modeling. Its mission is to provide a unified framework for writing various Arabic dialects, which differ significantly from Modern Standard Arabic (MSA) and lack standard orthography [16]. The goals and intentions of CODA* include maintaining internal consistency and coherence for writing DA, using the Arabic script, striking a balance between dialectal uniqueness and MSA-DA similarities, and ensuring ease of learning and readability [16]. The design principles of CODA* resemble English spelling in some ways, with phonological and historical considerations and some exceptions. It uses only Arabic script characters and can be written with or without diacritics [16]. Each DA word in CODA* has a unique orthographic form that represents its phonology, morphology, and meaning [16]. CODA* generally follows MSA-like orthographic decisions and preserves the phonological form of dialectal words, with some exceptions for highly variant root radical letters and pattern vowels [16]. It also maintains dialectal morphology and syntax, except for separating negation and indirect object pronouns. CODA* rules are dialect-independent and include specific rules for certain word classes to preserve morphological information, dialect integrity, and readability [16]. Overall, CODA* aims to make writing and learning dialectal Arabic more accessible and consistent. This is enhanced through the development of several tools to adjust the spelling of raw texts in Arabic languages to meet CODA* Guidelines such as the MADAR CODA Corpus [11]. Detailed instructions for the CODA* Guidelines are available at `https://camel-guidelines.readthedocs.io/`.

## 4 Proposed Approach

In our study, we aimed to improve the usability and clarity of the CODA* Guidelines by addressing user-reported issues. We engaged both native and non-native Tunisian speakers in structured interviews, live tests, and discussions on May 4 and May 24, 2017, considering the linguistic nuances of Tunisian Arabic. A text excerpt was dictated to three groups: A group provided with the CODA Guidelines as adapted for Tunisian by Zribi et al. (2014) [43], a group allowed to use the Tunisian Arabic Corpus (`https://www.tunisiya.org/`) by McNeil and Faiza [27] to see how to transcribe words, and a group with no support, spontaneously transcribing what was dictated. These tests were conducted by the Derja Association. Each group's output was assessed by another group to identify limitations. Structured interviews explored participants' experiences using common transcription rules for Tunisian. Choices to adjust the CODA* guidelines were made by a group of applied linguists based on the observations.

## 5 Differences from CODA* Guidelines

We will begin by establishing transcription conventions for consonants, then explore short and long-vowel transcription methods following NOTA Guidelines (Section 5.1). Finally, we will provide a comprehensive explanation of the NOTA orthography rules for Tunisian in the Arabic Script that do not apply to CODA* (Section 5.2). We provide the CODA* Guidelines that we kept for the NOTA Guidelines at `https://doi.org/10.6084/m9.figshare.25892974`.

### 5.1 Consonants and vowels

We follow the CODA* Guidelines for our revision [16] and introduce three supplemental letters from the Perso-Arabic Script to represent /p/ (پ), /v/

(پ), and /g/ (ڨ) [39]. These sounds are not adequately represented in Arabic, causing potential reading confusion [39]. Our additional letters align with common transcription practices among Tunisian speakers [39]. For the full table of consonant representations in Tunisian, refer to the table of consonants at https://doi.org/10.6084/m9.figshare.25892974.

As for the vowels, unlike Modern Standard Arabic, Tunisian Arabic features four vowel qualities (/a/, /ɛ/, /i/, and /u/) [20]. CODA* suggested representing long /ɛ/ as ِا (U+0650, U+0627) to distinguish it from long /a/ [16]. However, the use of ِا (U+0650, U+0627) might not be practical as it could potentially lead to confusion among Tunisian learners because ِ (U+0650) is mainly used to denote /i/. Furthermore, the rule for differentiating between /mˤ/, /nˤ/, /bˤ/, /zˤ/, and /rˤ/ and their non-emphatic minimal pairs also applies when the respective letter is adjacent to a short /a/. Therefore, a new Arabic diacritic for /ɛ/ needs to be defined. In this context, we propose adopting the Perso-Arabic diacritics used for Pashto, a language that also has four vowel qualities. We consequently use Fatha (U+064e), Zwarakai (U+0659), Damma (U+064f), and Kasra (U+0650) as diacritics. The details of vowel transcription for Tunisian are provided in the table of vowels at https://doi.org/10.6084/m9.figshare.25892974.

## 5.2 Revised Rules

Applying several CODA* guidelines to Tunisian revealed conflicts and user confusion. Thus, we adjusted several rules for a better transcription experience. First, the original CODA* Guidelines transcribe final /a/ and /ɛ/ as an Alif Maqsura ى, reflecting MSA etymology. For instance, /jɛ/ (arrive) is transcribed as جا because it appears in MSA as جاء [43,16]. We propose a morphology-aware rule: verbs are written with Alif Maqsura if the present form in the singular masculine third person ends with long /i/ ي. This applies to nouns derived from such verbs. For example, جاء in MSA becomes "جى [ʒæ]" in Tunisian because it conjugates as "يجي [jʒi]" in the present form of the singular masculine third person [6]. Another example is "ما كلموش [mæ kællmuːʃ]", meaning both *he does not call him* and "they did not call". To disambiguate, [uː] (meaning "him") can be replaced with هو [huː]. Second, morphological adjustments address irregularities in verb conjugation. In Tunis dialect, the imperative form of "كلا/klɛ/" is "كول [kuːl]" [6]. The second person singular present form "تاكل [tɛːkɪl]" suggests removing "ت" /t/ to form the imperative. Sfax dialect also uses "آكل [ʔɛːkɪl]" as the imperative, indicating the removal of exceptions could standardize Tunisian morphology, without favoring the Tunis dialect [18,12].

Third, we propose to disambiguate homonyms when possible. For example, "باش [bɛːʃ]" or "بش [bɪʃ]" combined with verbs means *will* (modal verb) [12]. We write it as "بِش [bəʃ]" to distinguish it from "باش [bɛːʃ]", meaning *with what*. Fourth, unlike the CODA* Guidelines, the definite article /il/ is always transcribed as ال (U+0627, U+0644) without diacritics to differentiate it from similar structures in diacritized texts. Fifth, we maintain a space between prepositions and the nouns they follow, despite Tunisian's agglutinative pattern. The same applies to the conjunction "و /w/" (U+0648), kept separate from the following noun. These choices were informed by examining the Tunisian Arabic Corpus [27] and user tests. Sixth, foreign vowels are transcribed according to their phonological similarity to Tunisian vowels, with [y] assimilated to /i/ and [ə] to /u/. Foreign vowels are written as long vowels unless in stressed syllables. Examples include loanwords like "فريكساي *fricassé*" (fried bread) and "أورديناتور *ordinateur*" (computer). Revised prepositions, foreign vowel conversion rules, and an example of how an excerpt is transcribed using spontaneous writing as well as NOTA and CODA* Guidelines are available at https://doi.org/10.6084/m9.figshare.25892974.

## 6 NOTA-Based Applications

The NOTA guidelines have potential across various domains, including education [21], language preservation [10], mass media [10], and cultural heritage [37]. However, the primary impact of this CODA* revision lies in fostering the development of linguistic resources and tools for Tunisian [41]. Current efforts to create language resources for Tunisian are limited to raw or annotated undiacritized corpora and bilingual lexicons [41], primarily supporting applications in computer science, including natural language processing and information retrieval such as sentiment analysis [24], machine translation [36], part-of-speech tagging [17], Arabic script transliteration of Latin script texts [25], Arabic diacritization [26], and morphological disambiguation [44]. Despite these efforts, Tunisian remains an under-resourced language, with text corpora containing only a few texts and lexicons with limited vocabulary [41]. The most extensive resource is the Tunisian Arabic Corpus, an online corpus with 2,462 texts totaling 1,085,493 words [27]. The revision of CODA*, through the NOTA guidelines, aims to address this limitation by enabling the development of large-scale corpora, lexicons, and resources for modeling Tunisian morphology.

### 6.1 Corpus and Lexicon Creation

Creating corpora and lexicons for Tunisian requires integrating linguistic expertise, computational methods, and ontological frameworks. This process is crucial for tasks like Arabic Diacritization and Named Entity Recognition (NER) [30]. Corpus creation involves compiling a representative collection of texts, annotated with linguistic information like diacritics and named entities [30]. Orthography guidelines ensure consistency in representing Tunisian Arabic, specifying how words and phrases, including diacritics, are written [43], enhancing corpus quality and supporting subsequent NLP tasks [30]. Orthography guidelines, particularly NOTA, are also essential for constructing well-organized lexicons. OntoLex and the Lexical Markup Framework (LMF) provide structured representations of lexical information, facilitating integration into semantic applications [14]. Adapting WordNet for Tunisian Arabic involves creating synsets and defining semantic relationships to represent the language's lexical nuances [30]. To incorporate Tunisian Arabic into multilingual knowledge graphs like Wikidata, specific Tunisian labels can be added using statements such as *rdfs:label* and *skos:altLabel*, establishing links to corresponding concepts and enabling cross-language information retrieval and knowledge sharing [40].

### 6.2 Morphology Modeling

Morphology modeling, crucial in natural language processing, studies the internal structure of words [7]. The Grammar Matrix is a framework enabling the construction of broad-coverage, cross-linguistically applicable grammars [7]. It allows customization by combining linguistic features [7]. The Grammatical Framework (GF) is another notable tool for morphology modeling [31], focusing on abstract syntax for generating sentences across languages [31]. The Arabic Treebank is significant for Arabic morphology modeling [22], providing annotated data for Arabic syntax and morphology, aiding computational model development [22]. While these initiatives support various languages, including Arabic, they have limited support for Arabic dialects like Tunisian. Efforts like CODA* have improved this situation, enabling the creation of small-scale treebanks for Tunisian [30]. NOTA guidelines aim to resolve conflicts in CODA*-based transcription of morphology rules.

## 7 Conclusion

In this research paper, we introduced the "Normalized Orthography for Tunisian Arabic" (NOTA), an adapted version of the CODA* guidelines tailored for transcribing Tunisian Arabic using the Arabic script. NOTA aims to bridge the gap between computational linguistics and wider acceptance of standardized Arabic script transcription for Tunisian Arabic. Through collaboration with native and non-native Tunisian speakers and extensive testing, NOTA emerges as a valuable resource for transcription and linguistic analysis, with future research focusing on its further refinement and utility in various applications.

**Acknowledgments.** We thank Mohamed Maamouri (University of Pennsylvania, United States of America), Nizar Habash (New York University Abu Dhabi, United Arab Emirates), Salam Khalifa (Stony Brook University, United States of America), Dominique Caubet (Alliance Sorbonne-Paris-Cité, France), Ramzi Hachani (United Nations Office for Project Services, United States of America), Ramzi Cherif (Derja Association, Tunisia), Rafik Zribi (IÉSEG School of Management, France), Lameen Souag (Alliance Sorbonne-Paris-Cité, France), Tariq Daouda (Université de Montréal,

Canada), and Nassim Regragui (Copenhagen Business School, Denmark) for useful comments and discussion. Certain portions of this work have been proofread using *ChatGPT*, a robust chatbot powered by OpenAI.

**Disclosure of Interests.** All the co-authors except Mohamed Ali Hadj Taieb and Mohamed Ben Aouicha are members of Derja Association, the non-profit organization promoting the widespread use of Tunisian.

# References

1. AlKhwiter, W., Al-Twairesh, N.: Part-of-speech tagging for Arabic tweets using CRF and Bi-LSTM. Computer Speech & Language **65**, 101138 (Jan 2021)
2. Alsudais, A., Alotaibi, W., Alomary, F.: Similarities between Arabic dialects: Investigating geographical proximity. Information Processing & Management **59**(1), 102770 (Jan 2022)
3. Althobaiti, M.J.: Creation of annotated country-level dialectal Arabic resources: An unsupervised approach. Natural Language Engineering **28**(5), 607–648 (Aug 2021)
4. Baccouche, T.: Le phonème /g/ dans les parlers arabes citadins de Tunisie. In: Proceedings of the seventh International Congress of Phonetic Sciences / Actes du Septième Congrès international des sciences phonétiques, pp. 652–655. De Gruyter (Dec 1972)
5. BBN Technologies, Makhoul, John, Zawaydeh, Bushra, Choi, Frederick, Stallard, David: BBN/AUB DARPA Babylon Levantine Arabic Speech and Transcripts. Linguistic Data Consortium (2005)
6. Ben Abdelkader, R.: Peace Corps English-Tunisian Arabic Dictionary. ERIC Clearinghouse (1977)
7. Bender, E.M., Flickinger, D., Oepen, S.: The grammar matrix: an open-source starter-kit for the rapid development of cross-linguistically consistent broad-coverage precision grammars. In: COLING-02 on Grammar Engineering and Evaluation. Association for Computational Linguistics (2002)
8. Besdouri, F.Z., Mekki, A., Zribi, I., Ellouze, M.: Improvement of the cota-orthography system through language modeling. In: 2021 IEEE/ACS 18th International Conference on Computer Systems and Applications (AICCSA). IEEE (Nov 2021)
9. Čéplö, S., Bátora, J., Benkato, A., Milička, J., Pereira, C., Zemánek, P.: Mutual intelligibility of spoken Maltese, Libyan Arabic, and Tunisian Arabic functionally tested: A pilot study. Folia Linguistica **50**(2) (Jan 2016)
10. Daoud, M.: The Language Situation in Tunisia. Current Issues in Language Planning **2**(1), 1–52 (Nov 2001)
11. Eryani, F., Habash, N., Bouamor, H., Khalifa, S.: A Spelling Correction Corpus for Multiple Arabic Dialects. In: Proceedings of the Twelfth Language Resources and Evaluation Conference. pp. 4130–4138. European Language Resources Association, Marseille, France (May 2020)
12. Gibson, M.: Tunis Arabic. Encyclopedia of Arabic language and linguistics **4**, 563–571 (2009)
13. Gibson, M.: Dialect Levelling in Tunisian Arabic: Towards a New Spoken Standard. In: Language Contact and Language Conflict in Arabic, pp. 42–58. Routledge (May 2013)

14. Gugliotta, E., Mallia, M., Panascì, L.: Towards a Unified Digital Resource for Tunisian Arabic. In: Carvalho, S., Khan, A.F., Anić, A.O., Spahiu, B., Gracia, J., McCrae, J.P., Gromann, D., Heinisch, B., Salgado, A. (eds.) Proceedings of the 4th Conference on Language, Data and Knowledge. pp. 579–590. NOVA CLUNL, Portugal, Vienna, Austria (Sep 2023)
15. Habash, N., Diab, M., Rambow, O.: Conventional Orthography for Dialectal Arabic. In: Proceedings of the Eighth International Conference on Language Resources and Evaluation (LREC'12). pp. 711–718. European Language Resources Association (ELRA), Istanbul, Turkey (May 2012)
16. Habash, N., Eryani, F., Khalifa, S., Rambow, O., et al.: Unified Guidelines and Resources for Arabic Dialect Orthography. In: Proceedings of the Eleventh International Conference on Language Resources and Evaluation (LREC 2018). European Language Resources Association (ELRA), Miyazaki, Japan (May 2018)
17. Hamdi, A., Nasr, A., Habash, N., Gala, N.: POS-tagging of Tunisian Dialect Using Standard Arabic Resources and Tools. In: Proceedings of the Second Workshop on Arabic Natural Language Processing. Association for Computational Linguistics (2015)
18. Kallel, M.A.: Choix langagiers sur la radio Mosaïque FM. Dispositifs d'invisibilité et de normalisation sociales. Langage et société **138**(4), 77–96 (Dec 2011)
19. Kassab, A., Ounaïes, A., Ben Achour Abdelkéfi, R., Louati, A., Mosbah, C., Sakli, M.: Histoire Générale de la Tunisie: Tome IV - L'Époque contemporaine. Sud Éditions (2010)
20. Maamouri, M.: The Phonology of Tunisian Arabic. Cornell University (1967)
21. Maamouri, M.: Language education and human development: Arabic diglossia and its impact on the quality of education in the Arab region. ERIC Clearinghouse (1998)
22. Maamouri, M., Bies, A., Buckwalter, T., Diab, M., et al.: Developing and Using a Pilot Dialectal Arabic Treebank. In: Proceedings of the Fifth International Conference on Language Resources and Evaluation (LREC'06). European Language Resources Association (ELRA), Genoa, Italy (May 2006)
23. Maamouri, M., Graff, D., Jin, H., Cieri, C., Buckwalter, T.: Dialectal Arabic Orthography-based Transcription & CTS Levantine Arabic Collection. the Parallel STT-NA Tracks Session of the EARS RT-04 Workshop, Palisades IBM Executive Center, New York, Nov. 10, 2004 (2004)
24. Masmoudi, A., Hamdi, J., Hadrich Belguith, L.: Deep Learning for Sentiment Analysis of Tunisian Dialect. Computación y Sistemas **25**(1) (Feb 2021)
25. Masmoudi, A., Khmekhem, M.E., Khrouf, M., Belguith, L.H.: Transliteration of Arabizi into Arabic Script for Tunisian Dialect. ACM Transactions on Asian and Low-Resource Language Information Processing **19**(2), 1–21 (Nov 2019)
26. Masmoudi, A., Mdhaffar, S., Sellami, R., Belguith, L.H.: Automatic Diacritics Restoration for Tunisian Dialect. ACM Transactions on Asian and Low-Resource Language Information Processing **18**(3), 1–18 (Jul 2019b)
27. McNeil, K.: Tunisian Arabic Corpus: Creating a Written Corpus of an 'Unwritten' Language, p. 30–55. Edinburgh University Press (2018)
28. McNeil, K.L.: Tunisian Arabic as a Written Language: Vernacularization and Identity. Georgetown University (2023)
29. Mekki, A., Zribi, I., Ellouze, M., Belguith, L.: Cota 2.0: an automatic corrector of tunisian arabic social media texts. Jordanian Journal of Computers and Information Technology **8**(4), 370–387 (2022)
30. Mekki, A., Zribi, I., Ellouze, M., Hadrich Belguith, L.: Critical description of TA linguistic resources. Procedia Computer Science **142**, 230–237 (2018)

31. Ranta, A.: Grammatical Framework. Journal of Functional Programming **14**(2), 145–189 (Jan 2004)
32. Rosa, C.L.: Mahdia Dialect: An Urban Vernacular in the Tunisian Sahel Context. Languages **6**(3), 145 (Aug 2021)
33. Rosso, P., Rangel, F., Farías, I.H., Cagnina, L., et al.: A survey on author profiling, deception, and irony detection for the Arabic language. Language and Linguistics Compass **12**(4) (Apr 2018)
34. Saddour, I.: The expression of progressivity in Tunisian Arabic: A study of progressive markers in oral retellings of simultaneous situations. Revue de Sémantique et Pragmatique **25-26**(Espace temps, Interprétations spatiales / Interprétations temporelles?), 265–280 (Apr 2009)
35. Sayahi, L.: Diglossia and Language Contact: Language variation and change in North Africa. Cambridge University Press (Apr 2014)
36. Sghaier, M.A., Zrigui, M.: Rule-Based Machine Translation from Tunisian Dialect to Modern Standard Arabic. Procedia Computer Science **176**, 310–319 (2020)
37. Shiri, S.: The Homestay in Intensive Language Study Abroad: Social Networks, Language Socialization, and Developing Intercultural Competence. Foreign Language Annals **48**(1), 5–25 (Mar 2015)
38. Singer, H.R.: Grammatik der arabischen Mundart der Medina von Tunis. DE GRUYTER (Jun 1984)
39. Turki, H., Adel, I., Daouda, T., Regragui, N.: A Conventional Orthography for Maghrebi Arabic. In: Proceedings of the International Conference on Language Resources and Evaluation (LREC) (2016)
40. Turki, H., Vrandecic, D., Hamdi, H., Adel, I.: Using WikiData as a Multi-lingual Multi-dialectal Dictionary for Arabic Dialects. In: 2017 IEEE/ACS 14th International Conference on Computer Systems and Applications (AICCSA). IEEE (Oct 2017)
41. Younes, J., Souissi, E., Achour, H., Ferchichi, A.: Language resources for Maghrebi Arabic dialects' NLP: a survey. Language Resources and Evaluation **54**(4), 1079–1142 (Apr 2020)
42. Zalmout, N., Habash, N.: Adversarial Multitask Learning for Joint Multi-Feature and Multi-Dialect Morphological Modeling. In: Proceedings of the 57th Annual Meeting of the Association for Computational Linguistics. Association for Computational Linguistics (2019)
43. Zribi, I., Boujelbane, R., Masmoudi, A., Ellouze, M., et al.: A Conventional Orthography for Tunisian Arabic. In: Proceedings of the Ninth International Conference on Language Resources and Evaluation (LREC'14). pp. 2355–2361. European Language Resources Association (ELRA), Reykjavik, Iceland (May 2014)
44. Zribi, I., Ellouze, M., Belguith, L.H., Blache, P.: Morphological disambiguation of Tunisian dialect. Journal of King Saud University - Computer and Information Sciences **29**(2), 147–155 (Apr 2017)
45. Zribi, I., Graja, M., Ellouze Khmekhem, M., Jaoua, M., Hadrich Belguith, L.: Orthographic Transcription for Spoken Tunisian Arabic, p. 153–163. Springer Berlin Heidelberg (2013)